\documentclass{article}
\usepackage{ijcai26}

\usepackage{times}
\usepackage{soul}
\usepackage{url}
\usepackage[utf8]{inputenc}
\usepackage[small]{caption}
\usepackage{graphicx}
\usepackage{amsmath}
\usepackage{booktabs}
\usepackage{algorithm}
\usepackage{algorithmic}
\usepackage[switch]{lineno}

\usepackage{pifont}
\usepackage{amsmath,amssymb,amsfonts}
\usepackage{textcomp}
\usepackage{multirow}
\usepackage{subfigure}
\usepackage{array}
\usepackage{colortbl}
\usepackage{xcolor}
\usepackage{mathrsfs}
\usepackage[colorlinks, citecolor=blue]{hyperref}

\title{Spatially Generalizable Mobile Manipulation via Adaptive Experience Selection and Dynamic Imagination}

\author{
Ping Zhong$^{1,2}$
\and
Liangbai Liu$^1$
\and
Bolei Chen$^{1}$\thanks{Corresponding Author.} 
\and
Tao Wu$^1$
\and \\
Jiazhi Xia$^{1,2}$
\and
Chaoxu Mu$^3$
\and
Jianxin Wang$^{1*}$
\affiliations
$^1$School of Computer Science and Engineering, Central South University\\
$^2$Xiangjiang Laboratory\\
$^3$School of Artificial Intelligence, Anhui University\\
\{liangbailiu, boleichen, 8102221321, ping.zhong, xiajiazhi\}@csu.edu.cn, cxmu@tju.edu.cn, jxwang@mail.csu.edu.cn
}

\begin{document}

\maketitle

\begin{abstract}

\textbf{M}obile \textbf{M}anipulation (MM) involves long-horizon decision-making over multi-stage compositions of heterogeneous skills, such as navigation and picking up objects. Despite recent progress, existing MM methods still face two key limitations: (i) low sample efficiency, due to ineffective use of redundant data generated during long-term MM interactions; and (ii) poor spatial generalization, as policies trained on specific tasks struggle to transfer to new spatial layouts without additional training. In this paper, we address these challenges through \textbf{A}daptive \textbf{E}xperience \textbf{S}election (AES) and model-based dynamic imagination. In particular, AES makes MM agents pay more attention to critical experience fragments in long trajectories that affect task success, improving skill chain learning and mitigating skill forgetting. Based on AES, a \textbf{R}ecurrent \textbf{S}tate-\textbf{S}pace \textbf{M}odel (RSSM) is introduced for \textbf{M}odel-\textbf{P}redictive \textbf{F}orward \textbf{P}lanning (MPFP) by capturing the coupled dynamics between the mobile base and the manipulator and imagining the dynamics of future manipulations. RSSM-based MPFP can reinforce MM skill learning on the current task while enabling effective generalization to new spatial layouts. Comparative studies across different experimental configurations demonstrate that our method significantly outperforms existing MM policies. Real-world experiments further validate the feasibility and practicality of our method. \href{https://307lab.github.io/SG-MM_Web/}{Project homepage}.

\end{abstract}

\section{Introduction}

\begin{figure}
    \centering
    \includegraphics[width=1\linewidth]{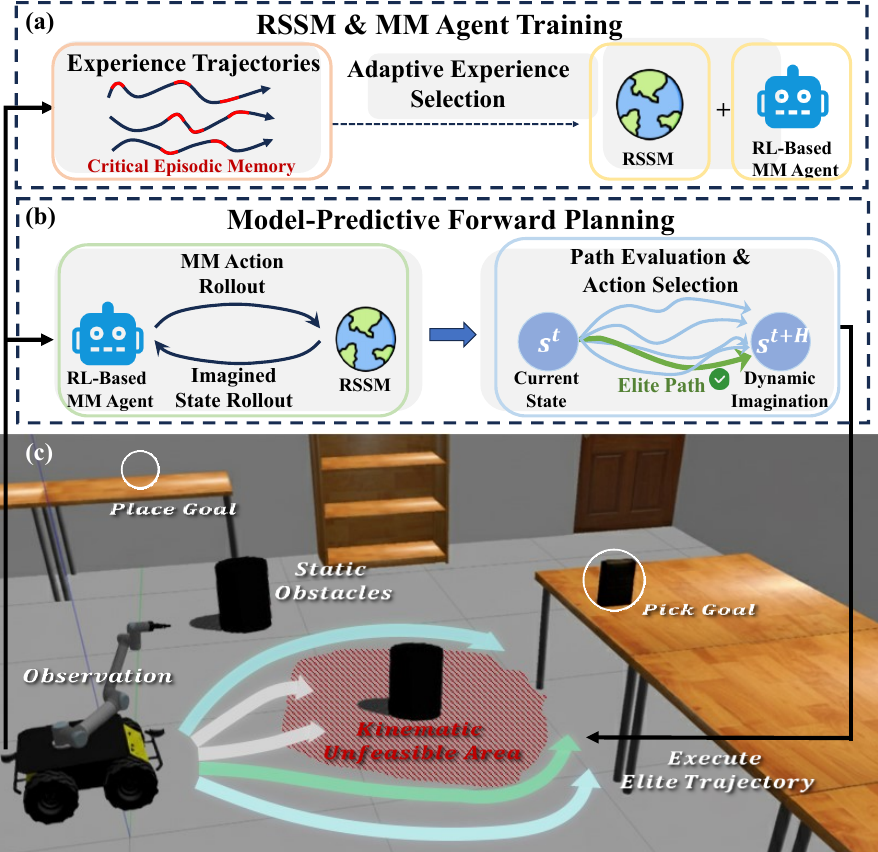}
    \caption{(a) Our AES enables RSSM to pay more attention to critical episodic memories in experience trajectories, which are used to improve and spatially generalize RL-based MM agents. (b) RSSM cooperates with RL-based MM skills for $H$-horizon dynamic imagination to achieve MPFP. The most promising elite path is selected for execution. (c) An example of MM tasks that consist of heterogeneous skills, i.e., navigation $\rightarrow$ picking $\rightarrow$ navigation $\rightarrow$ placing.}
    \label{fig1}
    \vspace{-0.5cm}
\end{figure}

\textbf{M}obile \textbf{M}anipulation (MM) is a cornerstone of embodied AI, enabling agents to perceive, reason, and act in unstructured environments. Unlike isolated navigation or fixed-base manipulation, MM requires agents to coordinate locomotion and dexterous interaction over extended spatial and temporal horizons. For example, MM allows embodied agents to complete complex goal-directed tasks such as opening doors and fetching objects, as shown in Fig. \ref{fig1} (c). MM is quite challenging, as it typically involves long-horizon decision-making over multi-stage compositions of heterogeneous skills \cite{huang2023skill,lin2025autoskill}. To address the complexity of MM, early methods decompose the problem into navigation \cite{chen2024embodied,2024HSPNav} and stationary manipulation \cite{nguyen2019review}. This factorized design simplifies modeling and learning, and has shown promise in structured environments. However, such approaches \cite{jauhri2024active,arduengo2021robust,feng2025predictive} often optimize only short-term objectives and overlook the dynamic coupling between mobility and manipulation, making them vulnerable to error accumulation over long horizons.

Recent work \cite{he2025flying,honerkamp2023n,huang2023skill} models the mobile base and manipulator as a unified system using whole-body control or joint policy learning. Despite encouraging progress, these solutions, especially model-free \textbf{R}einforcement \textbf{L}earning (RL)–based methods \cite{honerkamp2023n,huang2023skill}, usually suffer from inefficient exploration and low sample efficiency. The reason is that MM is essentially a long-horizon task consisting of skills with strict sequential dependencies, which introduces unique data distribution challenges that are different from those of standard RL benchmarking: \textbf{(1)} Failures in the early stages (navigation) can hinder the experience collection of all downstream tasks (picking and placing), thus hindering the learning of complete skill chains. \textbf{(2)} As the trajectory lengthens, the experience buffer becomes dominated by navigation-intensive fragments. This results in sparse but critical interaction moments in the trajectory being overwhelmed by redundant navigation experiences. \textbf{(3)} Such methods lack explicit mechanisms for learning reusable interaction patterns, which can easily lead to skill forgetting. Although recent data-driven or model-based methods \cite{chen2025ac,cen2025rynnvla,lin2025echovla} can help mitigate these issues, they are often sensitive to environmental variations and difficult to spatially generalize. The above limitations raise a fundamental question: \textbf{How can a MM policy efficiently exploit long-horizon experience to achieve robust decision-making and generalization across different spatial layouts?} 

Recent advances \cite{sun2023organizing,tafazoli2025building} in system neuroscience suggest that higher animals solve novel tasks by recalling task-relevant episodic fragments and composing them with previously acquired skills or habitual behaviors. Inspired by this, we aim to investigate how to memorize key and informative episodic fragments from rich experience trajectories, and use them to improve RL-learned MM policies so that they can adapt to new spatial layouts. To this end, we address the above limitations by recalling episodic memories and composing them with MM skills to produce robust and spatially generalizable MM behaviors.

As shown in Fig. \ref{fig1} (a), we propose \textbf{A}daptive \textbf{E}xperience \textbf{S}election (AES) to make MM agents pay more attention to key experience fragments in long trajectories that affect task success, improving skill chain learning and mitigating skill forgetting. Based on the informative state transitions well-selected by AES, a \textbf{R}ecurrent \textbf{S}tate-\textbf{S}pace \textbf{M}odel (RSSM) is introduced for \textbf{M}odel-\textbf{P}redictive \textbf{F}orward \textbf{P}lanning (MPFP) by capturing the coupled dynamics between the mobile base and the manipulator and imagining the dynamics of future manipulations. RSSM is trained together with model-free RL-based MM skill learning, and they share a single experience buffer. The difference is that AES helps RSSM memorize critical episodic fragments in a sample-efficient manner, thereby identifying weaknesses of the current MM skills and further consolidating them. As shown in Fig. \ref{fig1} (b), RSSM-based MPFP can prospectively imagine future state transitions and evaluate state goodness to support RL-based MM skill learning. As system neuroscience suggests, RSSM here acts as a function of memorizing and recalling episodic fragments, while RL-learned skills act as habitual behaviors specific to MM. Their combination can improve MM policies on the current task and generalize to new spatial layouts without performance loss and additional training. 

We implement our above ideas based on the popular \textbf{E}nd-\textbf{E}ffector (EE)-centric MM framework \cite{he2025flying,honerkamp2023n}. Our approach is evaluated through sufficient comparative studies in diverse experimental configurations, demonstrating significant performance gains over the \textbf{S}tate-\textbf{o}f-\textbf{T}he-\textbf{A}rt methods. Ablation studies further confirm the individual contributions of AES and RSSM-based MPFP to sample efficiency and spatial generalization. Finally, real-world robotic experiments highlight the practicality of our methods in realistic MM scenarios. In summary, our contributions are threefold: 

\textbf{(1)} We propose a system neuroscience-inspired method that leverages sample-efficient episodic recall to consolidate RL-based skill learning for robust and spatially generalizable MM. 

\textbf{(2)} An AES mechanism is introduced to make MM agents pay more attention to key experience fragments in long trajectories that affect task success, improving sample efficiency. 

\textbf{(3)} RSSM-based dynamic imagination is proposed for MPFP by capturing the coupled dynamics between the mobile base and the manipulator and imagining the dynamics of future manipulations.

\section{Related Work}

\textbf{Model-Free RL-based Mobile Manipulation.} Due to the difficulty of planning in the joint space of the mobile base and the manipulator, early work solves an MM task by decomposing it into sequential navigation and static manipulation tasks. Such a decomposition has been widely adopted in methods based on planning \cite{jauhri2024active,arduengo2021robust}, reachability analysis \cite{vahrenkamp2013robot,feng2025predictive}, and RL \cite{gu2022multi,huang2023skill,harish2024reinforcement,fang2022active}. Planning-based methods \cite{jauhri2024active,arduengo2021robust} plan trajectories for robots in joint space to ensure the kinematic feasibility of MM tasks. Given explicit task constraints, inverse reachability maps \cite{vahrenkamp2013robot} can also be exploited to determine favorable placements for the mobile base that facilitate subsequent manipulation. However, these classical methods are usually unable to instantly respond to diverse unstructured scenes and dynamic changes, and are thus less adaptable. Recently, RL-based approaches \cite{gu2022multi,huang2023skill,fang2022active} mitigate these issues by using hierarchical policy learning and curriculum learning techniques. In addition, there are methods \cite{jauhri2022robot} that use classical methods as behavioral priors to guide sample-efficient exploration. Most recently, some methods \cite{he2025flying,honerkamp2023n} investigate the EE-centric MM framework, which achieve strong autonomy and robustness by explicitly solving for base motion that cooperates with the manipulator during grasping.

Despite recent progress, existing MM policies still struggle with low sample efficiency, as they fail to exploit the sparse but informative transitions embedded in long-horizon trajectories. Moreover, policies learned for specific tasks generalize poorly to new spatial layouts due to the absence of mechanisms for memorizing and reusing task-relevant experience. In this work, we address these issues by proposing AES and RSSM-based dynamic imagination.

\textbf{Model-based Mobile Manipulation.} \textbf{V}ision-\textbf{L}anguage-\textbf{A}ction (VLA) models \cite{liu2025hybridvla,li2024cogact,zhang2025dreamvla} have recently emerged as a powerful paradigm for grounding multimodal instruction and perception into direct robot control. HybridVLA \cite{liu2025hybridvla} jointly integrates auto-regressive and diffusion-based policy heads to improve robustness across diverse manipulation tasks. EchoVLA \cite{lin2025echovla} augments VLA with declarative and episodic memory to better handle long-horizon MM and spatial context reasoning. Diffusion-based action models \cite{wen2025llada} introduce masked diffusion mechanisms tailored for structured action generation, setting new standards over auto-regressive VLAs in multimodal control. In parallel, world-model-augmented frameworks \cite{cen2025rynnvla,hafner2025dreamerv3} seek to unify dynamics prediction with action planning, enabling joint learning of environment dynamics and policy, which can improve long-horizon foresight. Despite these advances, current model-based methods often require large multimodal datasets and still struggle with sample efficiency and reliable generalization across diverse spatial layouts. In this work, we employ RSSM-based MPFP to improve and generalize RL-based MM skills to address these issues.

\section{Preliminaries} \label{sec3}

\textbf{Problem Definition.} The MM problem we consider in this work involves a robot that includes a mobile base and a manipulator. The MM tasks described in this paper involve skills such as navigation, collision avoidance, picking, placing, and opening. Such a problem is modeled as a goal-conditional \textbf{P}artially \textbf{O}bservable \textbf{M}arkov \textbf{D}ecision \textbf{P}rocess (POMDP) defined as a tuple $\mathcal{X}=(\mathcal{S}, \mathcal{O}, \mathcal{A}, \mathcal{P}, \mathcal{R}, \mathcal{G}, \rho, \gamma)$ with underlying state space $\mathcal{S}$, observation space $\mathcal{O}$, action space $\mathcal{A}$, reward function $\mathcal{R}:\mathcal{S}\times\mathcal{A} \to \mathbb{R}$, initial state distribution $\rho$ and discount factor $\gamma$. The state transition function $\mathcal{P}:\mathcal{S}\times\mathcal{A}\times\mathcal{S} \to \mathbb{R}_+$ represents the probability density of the next state $\boldsymbol{s}^{t+1} \in \mathcal{S}$ given the current state $\boldsymbol{s}^{t} \in \mathcal{S}$ and action $\boldsymbol{a}^{t} \in \mathcal{A}$. For tasks like object rearrangement \cite{habitatrearrangechallenge2022}, the goal space $\boldsymbol{g} \in \mathcal{G}$ is specified using the object's initial pickup pose or the goal pose where the object has to be placed. Our objective is to learn a policy $\pi_\theta(\boldsymbol{a}^t|\boldsymbol{o}^t,\boldsymbol{g})$ mapping an observation $\boldsymbol{o}^t \in \mathcal{O}$ and goal $\boldsymbol{g}$ to an action $\boldsymbol{a}^t$ that maximizes the expected return \cite{chen2023think}:
\setlength\abovedisplayskip{0.15cm}
\setlength\belowdisplayskip{0.15cm}
\begin{flalign} \small
\begin{aligned}
\pi_\theta^*(\boldsymbol{s}^t;\boldsymbol{g}) &= \arg\max_{\boldsymbol{a}^t} \Big[ \mathcal{R}(\boldsymbol{s}^t, \boldsymbol{a}^t;\boldsymbol{g}) + \\
&\gamma \int_{\boldsymbol{s}^{t+1}} \mathcal{P}(\boldsymbol{s}^{t+1}\mid \boldsymbol{s}^t,\boldsymbol{a}^t) V^*(\boldsymbol{s}^{t+1};\boldsymbol{g}) \, d\boldsymbol{s}^{t+1} \Big],
\end{aligned}
\label{eq1}
\end{flalign}
where $V^*(\boldsymbol{s}^t;\boldsymbol{g}) = \mathbb{E}_{\pi_\theta,\mathcal{P}}\Big[\sum_{k=t}^{T} \gamma^{k-t} \mathcal{R}(\boldsymbol{s}^k,\pi_\theta^*(\boldsymbol{s}^t;\boldsymbol{g}))\Big]$ is the optimal value function. In this setting, $\mathcal{P}(\boldsymbol{s}^{t+1}\mid \boldsymbol{s}^t,\boldsymbol{a}^t)$ represents the state transition over time, which is usually unknown to the MM agent. In this work, our MM policy recall key experience fragments and combine them with habitual skills to produce robust and spatially generalizable MM behaviors. Therefore, the state transition $\mathcal{P}$ and the optimal value function $V^*$ are learned together through dynamic imagination and model-free RL, respectively.

\begin{figure*}
    \centering
    \includegraphics[width=1.0\linewidth]{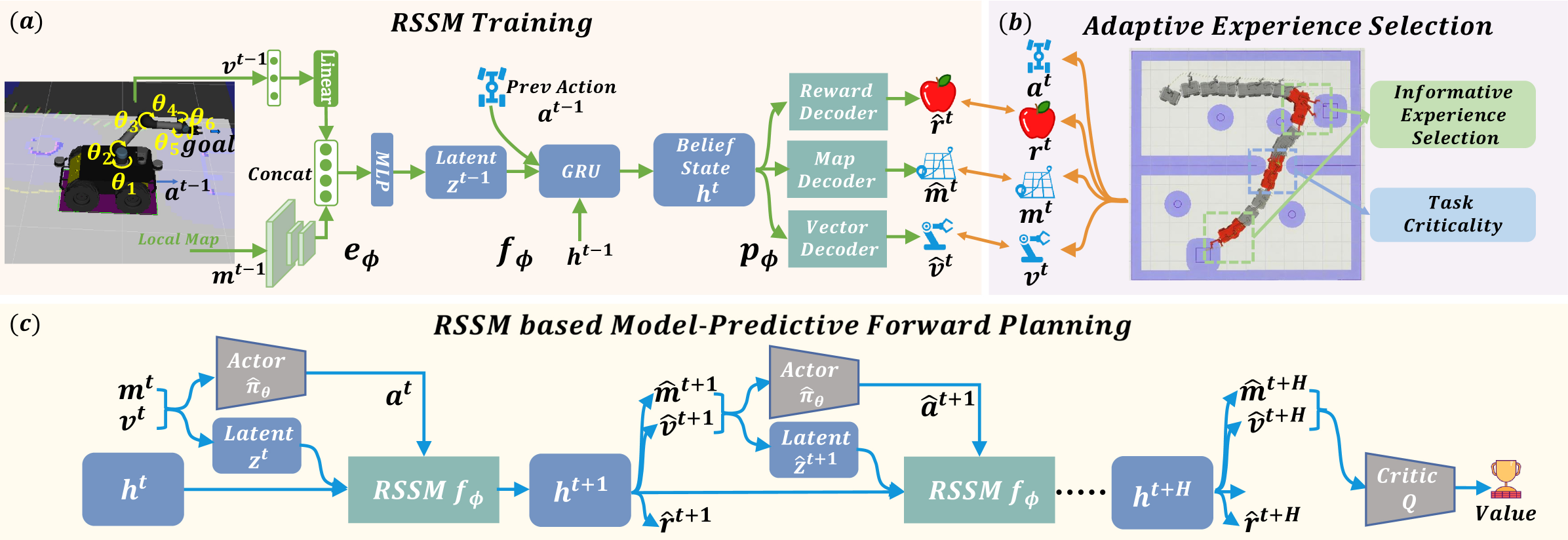}
    \vspace{-0.3cm}
    \caption{(a) An illustration of the structure, unfolding, and training process of RSSM. (b) An illustration of which key experience fragments the AES focuses on, including the experience of interacting with objects, opening doors, and passing through narrow areas. These experience segments are likely to have collisions and IK solving failures that affect task success. (c) An illustration of RSSM-based MPFP.}
    \label{fig2}
    \vspace{-0.5cm}
\end{figure*}

\textbf{EE-Centric MM.} To instantiate the MM tasks, we use the \textbf{R}obot \textbf{O}perating \textbf{S}ystem (ROS) to control an robotic system consisting of a Husky mobile base and a 6-DoF UR5 manipulator. Following the EE-centric MM framework \cite{he2025flying,honerkamp2023n}, given a sequence of 6D poses in \textit{SE}(3), the robot needs to find appropriate base placements in \textit{SE}(2) to allow the manipulator's EE to reach these 6D goal poses sequentially along dynamically feasible paths. Such a solution is natural, just as humans reach for objects through leg movements in conjunction with their arms. 

In practice, we employ an \textbf{I}nverse \textbf{K}inematics (IK) solver to solve the manipulator's motion in joint space. On this basis, our objective is to learn the navigation of the mobile base to cooperate with the long-horizon manipulation. This is challenging because the mobile base must move in a collision-free manner while ensuring that the next 6D goal pose of the manipulator is reachable. Otherwise, the MM task is terminated due to collision of the mobile base or IK solving failure of the manipulator. To ensure the robot's kinematic feasibility, the  mobile base's 2D action space $\mathcal{A} \in \mathbb{R}^2$ is continuous. The observation space $\mathcal{O}$ includes a robot-centric local occupancy map $\boldsymbol{m}^t$ and a state vector $\boldsymbol{v}^t$ consisting of the previous actions $\boldsymbol{a}^{t-1}$, the joint states of robot, the EE's pose and velocity, and the desired pose of EE. Such an EE-centric MM framework inherits decades of experience in the field of robotics and incorporates the advantages of robot learning while ensuring robustness and practicality.

\section{Methodology}

In the following section, we propose leveraging sample-efficient episodic recall to reinforce and spatially generalize RL-based MM skills. We first present how to employ AES-enhanced RSSM to memorize informative experience fragments for dynamic imagination (\S \ref{4.1}). Then, we present how to achieve MPFP based on RSSM to boost RL-based MM skill learning and promote spatial generalization (\S \ref{4.2}).

\subsection{AES-Enhanced RSSM for Dynamic Imagination} \label{4.1}

\textbf{Recurrent State-Space Model.} RSSM is responsible for modeling state transitions in a latent space to capture the coupled dynamics between mobile base and manipulator. We adopt a structured latent representation consisting of two components: the stochastic representation $\boldsymbol{z}^t$ and the recurrent state $\boldsymbol{h}^t$, maintaining consistent MM progress while incorporating new observations. An encoder $e_\phi$ is used to encode the observation $\boldsymbol{o}^{t-1}=\{\boldsymbol{m}^{t-1},\boldsymbol{v}^{t-1}\}$ as a stochastic representation $\boldsymbol{z}^{t-1}$, which is integrated into the recurrent state $\boldsymbol{h}^{t-1}$ to form a comprehensive state representation for MM decision-making. A recurrent module $f_\phi$ is utilized as the fundamental component of RSSM for latent state modeling, ensuring consistent temporal dynamics across state inference and future prediction. A decoder $p_\phi$ is used to predict observations and rewards from the recurrent state. As shown in Fig. \ref{fig2} (a), the components of our dynamic imagination model are outlined below:
\begin{flalign} \small
\begin{aligned}
\text{Observation Encoder:} & \boldsymbol{z}^{t-1}\ = e_\phi(\boldsymbol{o}^{t-1}), \\
\text{Dynamic Model: } & \boldsymbol{h}^{t} = f_\phi(\boldsymbol{h}^{t-1}, \boldsymbol{z}^{t-1},\boldsymbol{a}^{t-1}), \\
\text{Observation Predictor: } & \boldsymbol{\hat{o}^{t}} \sim p_\phi(\boldsymbol{\hat{o}^{t}|h^{t}}), \\
\text{Reward Predictor: } & \boldsymbol{\hat{r}}^{t} \sim p_\phi(\boldsymbol{\hat{r}}^{t}| \boldsymbol{h}^{t}).
\end{aligned}
\label{eq2}
\end{flalign}

According to the problem definition in Sec. \ref{sec3}, the dynamic model $f_\phi$ approximates $\mathcal{P}$ by modeling state transitions. The training of $e_\phi$, $f_\phi$, and $q_\phi$ is parallel to RL-based MM skill learning, and they share the same experience buffer. The difference is that RSSM's training data is well-selected by AES as follows. The loss functions used to train RSSM are consistent with Dreamer V3 \cite{hafner2025dreamerv3}.

\textbf{Adaptive Experience Selection.} MM skills learned in sample-inefficient manner are difficult to generalize to new spatial layouts. On the one hand, standard uniform sampling is insufficient, as it tends to overfit the abundant navigation data while neglecting the critical manipulation experience strongly associated with task success. On the other hand, focusing solely on hard failures risks catastrophic forgetting of foundational navigation capabilities. To address this challenge, we introduce an AES mechanism that pays more attention to critical state transitions in MM trajectories, as shown in Fig. \ref{fig2} (b). We aim to maintain a balance between common navigation and critical failure experience, thus fostering a dynamic model that is robust across all skills and diverse spatial layouts to enhance generalization and prevent skill forgetting. Technically, the RSSM is trained on fixed-length experience fragments that are extracted from episodes collected by the RL-based MM agent. For each MM episode that is put into the experience buffer, the experience segments are sliding experience windows that contain consecutive state transitions that have the highest total prioritization. Prioritization is achieved by assigning the $i$-th state transition a composite priority score $P^i_{total}$, which integrates three complementary signals:

\textbf{(1) Informative Experience Selection:} From a data distribution perspective, informative state transitions that exhibit significant changes usually account for a small percentage and can thus be considered outliers. These state transitions usually correspond to moments such as opening a door, picking up or placing the target object, and avoiding collisions, as shown in Fig. \ref{fig2} (b). Therefore, we propose to transform the task of selecting critical experience into the detection of outliers in the feature space of observations. Specifically, given a sequence of observations $\{\boldsymbol{o}^1,\boldsymbol{o}^2,...,\boldsymbol{o}^T\}$, where $\boldsymbol{o}^t = \{\boldsymbol{m}^t,\boldsymbol{v}^t\}$, we use $d(\boldsymbol{o}^i,\boldsymbol{o}^j)$ to denote the distance in feature space:
\begin{equation} \small
d(\boldsymbol{o}^i,\boldsymbol{o}^j)=||\boldsymbol{m}^j-\boldsymbol{m}^i||_1 + \lambda \cdot (1-\frac{\boldsymbol{v}^{i\top}\boldsymbol{v}^j}{||\boldsymbol{v}^i||_2\cdot||\boldsymbol{v}^j||_2}),
\label{eq3}
\end{equation}
where $\lambda$ is a weight used to balance magnitude. Based on Eq. (\ref{eq3}), we define the $k$-distance neighborhood: $N_k(\boldsymbol{o}^i)=\{\boldsymbol{o}^j | d(\boldsymbol{o}^i,\boldsymbol{o}^j) \le d_k(\boldsymbol{o}^i)\}$, where $d_k(\boldsymbol{o}^i)$ is the distance to the $k$-th nearest neighbor. For each neighbor $\boldsymbol{o}^j \in N_k(\boldsymbol{o}^i)$, the reachability distance is defined as $\text{reach-}d_k(\boldsymbol{o}^i,\boldsymbol{o}^j)=max\{d_k(\boldsymbol{o}^j),d(\boldsymbol{o}^i,\boldsymbol{o}^j)\}$. Using this, the \textbf{L}ocal \textbf{O}utlier \textbf{F}actor (LOF) \cite{breunig2000lof} of $\boldsymbol{o}^i$ is the inverse of its mean reachability distance:
\begin{equation} \small
\text{LRD}_k(\boldsymbol{o}^i) = \left( \frac{1}{|N_k(\boldsymbol{o}^i)|} \sum_{\boldsymbol{o}^j \in N_k(\boldsymbol{o}^i)} \text{reach-}d_k(\boldsymbol{o}^i, \boldsymbol{o}^j) \right)^{-1}.
\label{eq4}
\end{equation}
Then, the LOF is calculated as the relative density of $\boldsymbol{o}^i$ compared to its neighbors:
\begin{equation} \small
P^i_{ice} = \text{LOF}_k(\boldsymbol{o}^i) = \frac{1}{|N_k(\boldsymbol{o}^i)|} \sum_{\boldsymbol{o}^j \in N_k(\boldsymbol{o}^i)} \frac{\text{LRD}_k(\boldsymbol{o}^j)}{\text{LRD}_k(\boldsymbol{o}^i)} 
\label{eq5}
\end{equation}
Based on this formulation, the observation is selected when $\text{LOF}_k(\boldsymbol{o}^i)>1$. A high LOF$_k(\boldsymbol{o}^i)$ indicates a novel and unusual state, receiving proportional priority.

\textbf{(2) Task Criticality:} This signal marks state transitions containing IK solving failures of the manipulator. These informative failure modes receive significant prioritization $P_{tc}$, ensuring that RSSM consistently focuses on learning to navigate to coordinate with successful IK solving, thereby preventing future failures.

\textbf{(3) Prediction Error:} We follow the idea of prioritized experience replay \cite{schaul2015prioritized}, employing the RSSM's overall training loss as a dynamic indicator of learning difficulty. The state transitions that are modeled incorrectly by RSSM are given higher priorities $P_{per}$, which the RSSM should focus on in the following training.

Overall, the final priority $P_{total}^i$ for $i$-th state transition is computed as a weighted sum of three signals:
\begin{equation} \small
\begin{split}
P_{total}^i = w_1 \max\bigl(0,\, P_{ice}^i - 1\bigr) + 
w_2P_{tc}^i \cdot \mathbb{I} + w_3P_{per}^i,
\end{split}
\label{eq6}
\end{equation}
where $w_1$, $w_2$, $w_3$ are respective weights and $\mathbb{I}$ indicates whether an IK solving failure occurs. In practice, all high-priority experience fragments are structured as a SumTree \cite{wang2024prioritized} to efficiently manage and sample them based on normalized prioritization. 


\vspace{-0.2cm}
\subsection{Dynamic Imagination-based Foresighted MM} \label{4.2}


Since RSSM and MM skills are trained together, AES-enhanced RSSM can provide foresighted guidance for RL-based MM skill learning through dynamic imagination. In addition, we emphasize using the AES-enhanced RSSM to generalize fully trained MM agents to new spatial layouts without additional learning. Following common practice \cite{jauhri2022robot,honerkamp2021learning}, we model a maximum-entropy MM policy $\hat{\pi}_\theta$ to explore the state space of a given task using the \textbf{S}oft \textbf{A}ctor-\textbf{C}ritic (SAC) algorithm \cite{haarnoja2018soft}. The following RSSM-based MPFP is introduced to reinforce and spatially generalize MM skills during training and evaluation, respectively.

\begin{figure*}[t]
 \centering
 \includegraphics[width=1.0\linewidth]{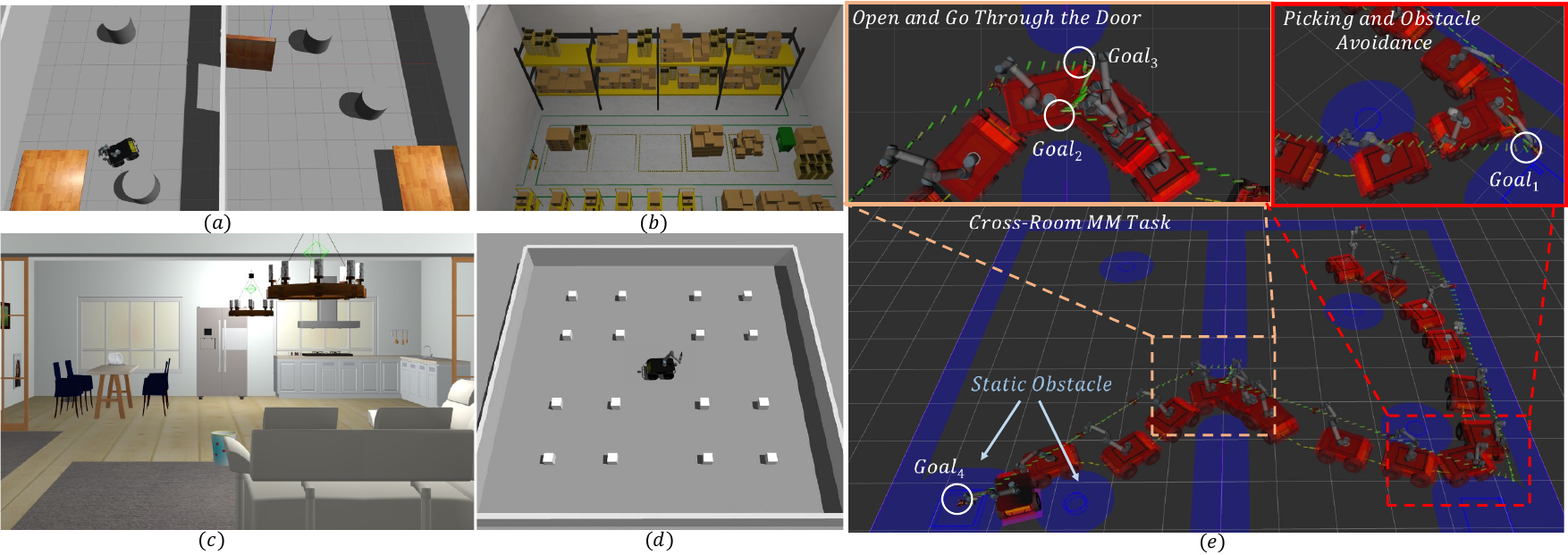}
  \vspace{-0.3cm}
 \caption{We consider four different experimental configurations: (a) cross-room, (b) warehouse-oriented, (c) home-scene-oriented, and (d) dynamic-scene-oriented (white dynamic obstacles) mobile manipulations. (e) An illustration of the cross-scene MM task. In each episode, the positions of the robot, the obstacles, the desks, the picking point (Goal$_1$), the start point of opening the door (Goal$_2$), the end point of opening the door (Goal$_3$), and the placing point (Goal$_4$) are randomly initialized. This task requires the robot to pick up an object and then open a door to another room to place the object in a specified position in the room. More details can be found in the supplementary material.}
 \label{fig3}
 \vspace{-0.5cm}
\end{figure*}

\textbf{Model-Predictive Forward Planning.} During MM decision-making in both training and evaluation, the agent does not directly execute the action predicted by the policy $\hat{\pi}_\theta$ learned by RL. Instead, it leverages RSSM as an internal simulator for dynamic imagination. As shown in Fig. \ref{fig2} (c), this process integrates the concurrently trained actor-critic networks $(\hat{\pi}_\theta, Q)$, which are used for the action rollout in dynamic imagination and the value estimation of the last imagined step, respectively. At each timestep, the MM agent samples action sequences from a Gaussian distribution $\mathcal{N}(\mu,\sigma)$ whose initial parameters are initialized from the actor $\hat{\pi}_\theta(\boldsymbol{o}^t;\boldsymbol{g})$ based on observation $\boldsymbol{o}^t$ and goal $\boldsymbol{g}$. During each iteration of \textbf{C}ross-\textbf{E}ntropy \textbf{M}ethod (CEM) \cite{rubinstein1999cross}, $L$ candidate action sequences of horizon $H$ are drawn and evaluated in parallel by unrolling the RSSM:
\begin{flalign} \small
\begin{aligned}
\text{Dynamic Imagination: } & \boldsymbol{h}^{k+1}_i = f_\phi(\boldsymbol{h}^{k}_i, \boldsymbol{z}^{k}_i,\boldsymbol{a}^{k}_i), \\
\text{Observation Rollout: } & \boldsymbol{\hat{o}}^{k+1}_i \sim p_\phi(\boldsymbol{\hat{o}}^k_i|\boldsymbol{\hat{h}}^k_i),\\
\text{Action Rollout: } & \boldsymbol{\hat{a}}^{k+1}_i = \hat{\pi}_\theta(\boldsymbol{\hat{o}}^k_i;\boldsymbol{g}),\\
\text{Observation Encoding: } & \boldsymbol{\hat{z}}^{k+1}_i = e_\phi(\boldsymbol{\hat{o}}^k_i),\\
\text{Reward Prediction: } & \boldsymbol{\hat{r}}^{k+1}_i \sim p_\phi(\boldsymbol{\hat{r}}^{k+1}_i| \boldsymbol{h}^{k+1}_i),
\end{aligned}
\label{eq7}
\end{flalign}
where $i=0,...,L-1$ and $k \in [t, t+H-1]$. As shown in Fig. \ref{fig2} (c), each $H$-horizon dynamic imagination yields a cumulative return:
\begin{flalign} \small
\begin{aligned}
\hat{\mathcal{R}}^t =  \sum_{k=t}^{t+H-1} \gamma^{k-t}\hat{\boldsymbol{r}}^{k}_i
    + \gamma^{t+H} Q(\boldsymbol{\hat{o}}^{t+H}_i,\hat{\pi}_\theta(\boldsymbol{\hat{o}}^{t+H}_i;\boldsymbol{g})),
\end{aligned}
\label{eq8}
\end{flalign}
which is an estimate of
\[
\gamma \int_{\boldsymbol{s}^{t+1}} \mathcal{P}(\boldsymbol{s}^{t+1}\mid \boldsymbol{\hat{o}}^t,\boldsymbol{a}^t) V^*(\boldsymbol{s}^{t+1};g) \, d\boldsymbol{s}^{t+1}
\]
in Eq. (\ref{eq1}). Notably, the critic $Q$ provides a terminal value estimate based on the last imagined observation $\boldsymbol{\hat{o}}^{t+H}_i$, which is decoded from the belief state $\boldsymbol{h}^{t+H}_i$. The top-$K$ sequences with the highest returns are selected as elites, and the distribution parameters $(\mu,\sigma)$ of $\hat{\pi}_\theta$ are refitted to their empirical statistics. This sampling-evaluation-refitting loop repeats for a fixed number of iterations. After convergence, only the first action of the final mean sequence is executed by the MM agent, and the overall process repeats at the next timestep, forming a standard model-predictive cycle. The combination of RSSM rollouts, MM decision-making, and CEM optimization enables robust and foresighted planning under complex dynamics and uncertainty.

\section{Experiments}

Based on the EE-centric MM framework introduced in Sec. \ref{sec3}, we answer the following research questions through comparative and ablative studies: \textbf{(1)} How does our method perform compared to strong baselines? \textbf{(2)} Does RSSM-based MPFP help to generalize the model-free RL-based MM skills to new spatial layouts? \textbf{(3)} Does our proposed AES mechanism help improve sample efficiency? \textbf{(4)} Does RSSM-based MPFP help to reinforce the MM policy? \textbf{(5)} How well does our MM policy perform on the real-world robot?

\subsection{Experimental Setup}

\textbf{Experimental Configurations.} As shown in Fig. \ref{fig3}, we follow the experimental configurations in N$^2$M$^2$ \cite{honerkamp2021learning} and consider four different experimental configurations: cross-room, warehouse-oriented, home-scene-oriented, and dynamic-scene-oriented mobile manipulations. We first train and evaluate the proposed method and baselines individually in each of the four scenes. Since cross-room MM involves more diverse skills, including navigation, collision avoidance, picking up, opening doors, and placing, we evaluate the performance of generalizing the MM agent fully trained in such a configuration to the other three configurations. This refers to generalizing MM agents to new spatial layouts without additional training. Please see the supplementary material for more details.

\textbf{Baselines and Evaluation Metrics.} We compare our proposed method with several model-free RL-based MM policies, including the SoTA EE-centric MM method N$^2$M$^2$ \cite{honerkamp2021learning}, an end-to-end whole-body control method adapted from N$^2$M$^2$, the award-winning hierarchical method BHyRL \cite{jauhri2022robot}, and the SoTA approach based on auxiliary task distillation \cite{harish2024reinforcement}. In addition, we compare our method with two world model-based MM policies: \textbf{(1) Dreamer V3-based MM} \cite{hafner2025dreamerv3}: This baseline adapts the Dreamer V3 architecture to the MM setting by learning a latent world model and performing latent imagination rollout for policy optimization. We modify the open-source implementation of Dreamer V3 to make it suitable for EE-centric MM to obtain experimental results. \textbf{(2) TD-MPC2-based MM} \cite{hansen2024tdmpc2}: This baseline applies TD-MPC2 to the EE-centric MM setting as a model-based MPC-style planner that uses online trajectory optimization in the latent state space. Similar to our approach, only the first action from the optimized sequence is executed in the environment, resulting in a receding-horizon control loop suitable for complex MM dynamics.


We report the \textbf{A}verage \textbf{IK} solution \textbf{F}ailures (AIKF), \textbf{A}verage \textbf{B}ase \textbf{C}ollisions (ABC), \textbf{T}ask \textbf{C}ompletion \textbf{R}ate (TCR), and \textbf{P}erfect \textbf{S}uccess \textbf{R}ate (PSR) as evaluation metrics. A lower AIKF means a better cooperation between the mobile base and the manipulator. TCR implies that the MM robotic system completes the task while allowing for IK solving failures and base collisions. We set the threshold for allowing both kinds of failures to 20. PSR means completing the MM tasks with zero failures.

\renewcommand\arraystretch{1.0}
\begin{table}[t] \small
\caption{Comparative studies are conducted in the cross-room experimental configuration to demonstrate the superiority of our method. Each method was evaluated over 100 episodes across 3 random seeds. We report the mean and standard deviation of each metric.}
\vspace{-0.3cm}
\begin{center}
\setlength{\tabcolsep}{2.0mm}{
\begin{tabular}{l c c c c }
\hline
\multirow{2}{*}{\textbf{Method}} &
 \multicolumn{4}{c}{\textbf{Cross-room}} \\
\cline{2-5}
 & \textbf{AIKF}$\downarrow$ & \textbf{ABC}$\downarrow$ & \textbf{TCR}$\uparrow$ & \textbf{PSR}$\uparrow$  \\
\hline
End-to-End MM & - & - & 0$_{\textcolor{gray}{\pm 0}}$ & 0$_{\textcolor{gray}{\pm 0}}$ \\
BHyRL & 12.1$_{\textcolor{gray}{\pm 1.2}}$ & 2.8$_{\textcolor{gray}{\pm 0.5}}$ & 21$_{\textcolor{gray}{\pm 3}}$ & 19$_{\textcolor{gray}{\pm 4}}$  \\
N$^2$M$^2$ & 11.3$_{\textcolor{gray}{\pm 1.3}}$ & 2.2$_{\textcolor{gray}{\pm 0.6}}$ & 29$_{\textcolor{gray}{\pm 4}}$ & 17$_{\textcolor{gray}{\pm 2}}$ \\
Dreamer V3 & 8.1$_{\textcolor{gray}{\pm 1.3}}$ & 2.0$_{\textcolor{gray}{\pm 0.5}}$ & 54$_{\textcolor{gray}{\pm 4}}$ & 40$_{\textcolor{gray}{\pm 1}}$ \\
TD-MPC2 & 9.9$_{\textcolor{gray}{\pm 1.6}}$ & \textbf{1.2}$_{\textcolor{gray}{\pm 0.4}}$ & 46$_{\textcolor{gray}{\pm 3}}$ & 41$_{\textcolor{gray}{\pm 1}}$ \\
AuxDistill & 8.8$_{\textcolor{gray}{\pm 1.5}}$ & 1.3$_{\textcolor{gray}{\pm 0.3}}$ & 57$_{\textcolor{gray}{\pm 6}}$ & 42$_{\textcolor{gray}{\pm 7}}$ \\
\hline
Ours & \textbf{4.1}$_{\textcolor{gray}{\pm 1.1}}$ & 2.0$_{\textcolor{gray}{\pm 0.2}}$ & \textbf{81}$_{\textcolor{gray}{\pm 3}}$ & \textbf{54}$_{\textcolor{gray}{\pm 2}}$ \\
\hline
\end{tabular}}
\end{center}
\vspace{-0.5cm}
\label{table1}
\end{table}

\textbf{Implementation Details.} We use a 3-layer CNN with maximal pooling to encode the local occupancy map, whose size and resolution are 30 $\times$ 30 and 0.1 m, respectively. Other observations, e.g., the state of the robot, are concatenated with the map embeddings and then passed through a multilayer perceptron. In our experiments, the RL agents are trained for $5 \times 10^{6}$ steps of environmental interactions. The learning rate is linearly decayed from an initial value of $1 \times 10^{-4}$ to a final value of $1 \times 10^{-6}$ over the course of training. The reward discount factor $\gamma$ is set to 0.99. For RSSM training, the loss weights $w_1$, $w_2$, and $w_3$ in Eq. (\ref{eq6}) are all set to 1.0. We employ a linear schedule for the planning horizon $H$, linearly increasing it from 1 to 8 over $5 \times 10^{6}$ steps. During evaluation, $H$ is set to 8. The number of candidate action sequences $L$ is set to 50. The Top-10 sequences are selected as elites during CEM optimization.

\begin{table*} \small
\caption{Comparative studies are conducted in three other experimental configurations to demonstrate the superiority of our method.}
 \vspace{-0.3cm}
\begin{center}
\setlength{\tabcolsep}{2.0mm}{
\begin{tabular}{l c c c c | c c c c | c c c c }
\hline
\multirow{2}{*}{\textbf{Method}} &
 \multicolumn{4}{c}{\textbf{Home-scene}} &
 \multicolumn{4}{c}{\textbf{Warehouse}} &
 \multicolumn{4}{c}{\textbf{Dynamic-scene}}\\
\cline{2-13}
 & \textbf{AIKF}$\downarrow$ & \textbf{ABC}$\downarrow$ & \textbf{TCR}$\uparrow$ & \textbf{PSR}$\uparrow$ & \textbf{AIKF}$\downarrow$ & \textbf{ABC}$\downarrow$ & \textbf{TCR}$\uparrow$ & \textbf{PSR}$\uparrow$ & \textbf{AIKF}$\downarrow$ & \textbf{ABC}$\downarrow$ & \textbf{TCR}$\uparrow$ & \textbf{PSR}$\uparrow$  \\
\hline
N$^2$M$^2$ & 11.0$_{\textcolor{gray}{\pm 0.9}}$ & 2.3$_{\textcolor{gray}{\pm 0.5}}$ & 32$_{\textcolor{gray}{\pm 4}}$ & 18$_{\textcolor{gray}{\pm 2}}$ & 7.1$_{\textcolor{gray}{\pm 1.5}}$ & 0.0$_{\textcolor{gray}{\pm 0.0}}$ & 68$_{\textcolor{gray}{\pm 2}}$ & 56$_{\textcolor{gray}{\pm 3}}$ & 16.5$_{\textcolor{gray}{\pm 1.1}}$ & 6.4$_{\textcolor{gray}{\pm 0.8}}$ & 12$_{\textcolor{gray}{\pm 2}}$ & 3$_{\textcolor{gray}{\pm 1}}$\\
Dreamer V3 & 9.3$_{\textcolor{gray}{\pm 0.7}}$ & 1.6$_{\textcolor{gray}{\pm 0.5}}$ & 64$_{\textcolor{gray}{\pm 2}}$ & 45$_{\textcolor{gray}{\pm 2}}$ & 5.3$_{\textcolor{gray}{\pm 1.2}}$ & 0.4$_{\textcolor{gray}{\pm 0.2}}$ & 80$_{\textcolor{gray}{\pm 3}}$ & 61$_{\textcolor{gray}{\pm 1}}$ & 9.0$_{\textcolor{gray}{\pm 0.9}}$ & \textbf{3.1}$_{\textcolor{gray}{\pm 0.4}}$ & 44$_{\textcolor{gray}{\pm 3}}$ & \textbf{40}$_{\textcolor{gray}{\pm 3}}$\\
TD-MPC2 & 10.7$_{\textcolor{gray}{\pm 1.1}}$ & 1.5$_{\textcolor{gray}{\pm 0.4}}$ & 56$_{\textcolor{gray}{\pm 3}}$ & 32$_{\textcolor{gray}{\pm 2}}$ & 6.0$_{\textcolor{gray}{\pm 1.0}}$ & 0.5$_{\textcolor{gray}{\pm 0.5}}$ & 74$_{\textcolor{gray}{\pm 2}}$ & 60$_{\textcolor{gray}{\pm 1}}$ & 8.9$_{\textcolor{gray}{\pm 1.0}}$ & 5.0$_{\textcolor{gray}{\pm 0.6}}$ & 40$_{\textcolor{gray}{\pm 2}}$ & 35$_{\textcolor{gray}{\pm 1}}$\\
\hline
Ours & \textbf{3.6}$_{\textcolor{gray}{\pm 1.1}}$ & \textbf{0.1}$_{\textcolor{gray}{\pm 0.1}}$ & \textbf{90}$_{\textcolor{gray}{\pm 2}}$ & \textbf{66}$_{\textcolor{gray}{\pm 4}}$ & \textbf{3.0}$_{\textcolor{gray}{\pm 1.0}}$ & \textbf{0.0}$_{\textcolor{gray}{\pm 0.0}}$ & \textbf{96}$_{\textcolor{gray}{\pm 4}}$ & \textbf{76}$_{\textcolor{gray}{\pm 3}}$ & \textbf{5.8}$_{\textcolor{gray}{\pm 1.6}}$ & 5.6$_{\textcolor{gray}{\pm 0.4}}$ & \textbf{58}$_{\textcolor{gray}{\pm 3}}$ & 39$_{\textcolor{gray}{\pm 1}}$\\
\hline
\end{tabular}}
\end{center}
\label{table2}
\vspace{-0.3cm}
\end{table*}

\begin{figure}[t]
 \centering
 \includegraphics[width=0.9\linewidth]{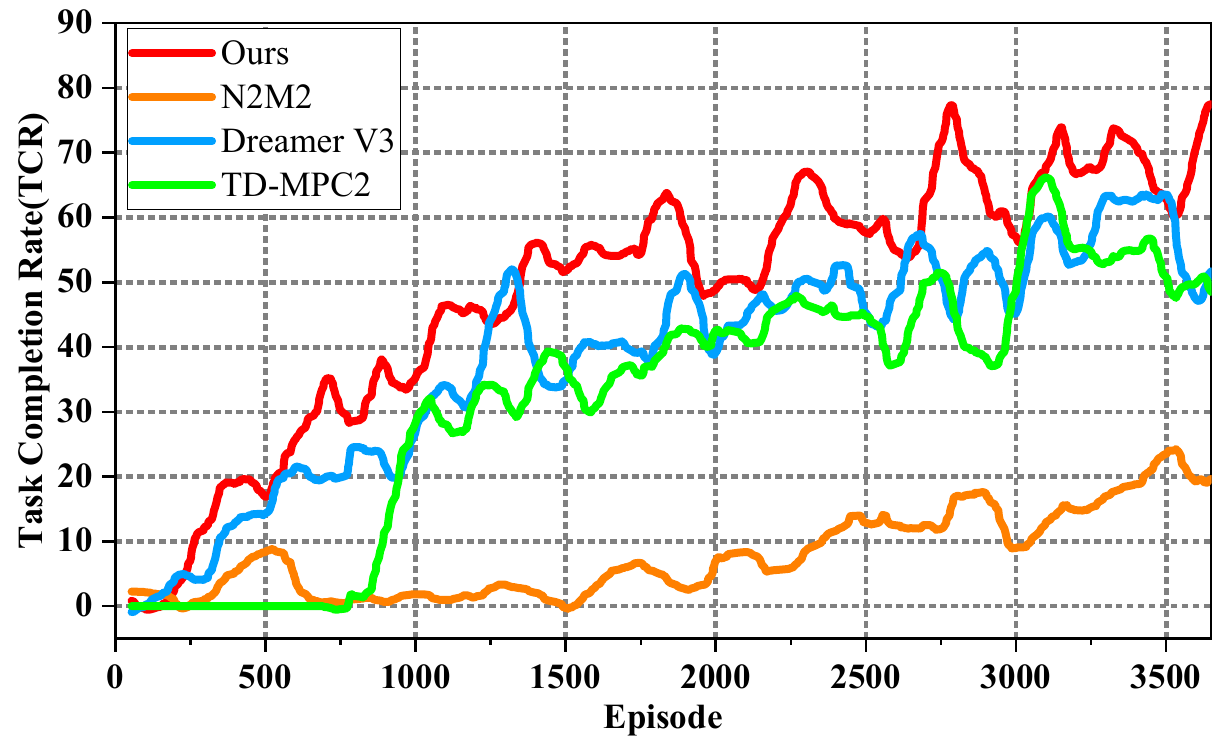}
  \vspace{-0.3cm}
 \caption{TCR metrics change with the number of training episodes in the cross-room experimental configuration.}
 \label{fig4}
 \vspace{-0.5cm}
\end{figure}

\begin{table*} \small
\caption{The models trained in the cross-room experimental configuration are transferred to three new experimental configurations for evaluation without additional training.}
 \vspace{-0.3cm}
\begin{center}
\setlength{\tabcolsep}{1.6mm}{
\begin{tabular}{l c c c c | c c c c | c c c c }
\hline
\multirow{2}{*}{\textbf{Method}} &
 \multicolumn{4}{c}{\textbf{Home-scene}} &
 \multicolumn{4}{c}{\textbf{Warehouse}} &
 \multicolumn{4}{c}{\textbf{Dynamic-scene}}\\
\cline{2-13}
& \textbf{AIKF}$\downarrow$ & \textbf{ABC}$\downarrow$ & \textbf{TCR}$\uparrow$ & \textbf{PSR}$\uparrow$ & \textbf{AIKF}$\downarrow$ & \textbf{ABC}$\downarrow$ & \textbf{TCR}$\uparrow$ & \textbf{PSR}$\uparrow$ & \textbf{AIKF}$\downarrow$ & \textbf{ABC}$\downarrow$ & \textbf{TCR}$\uparrow$ & \textbf{PSR}$\uparrow$  \\
\hline
N$^2$M$^2$ & 15.1$_{\textcolor{gray}{\pm 1.3}}$ & 1.4$_{\textcolor{gray}{\pm 0.3}}$ & 29$_{\textcolor{gray}{\pm 1}}$ & 15$_{\textcolor{gray}{\pm 1}}$ & 14.3$_{\textcolor{gray}{\pm 1.2}}$ & 0.0$_{\textcolor{gray}{\pm 0.0}}$ & 19$_{\textcolor{gray}{\pm 1}}$ & 11$_{\textcolor{gray}{\pm 1}}$ & 18.7$_{\textcolor{gray}{\pm 1.6}}$ & 4.1$_{\textcolor{gray}{\pm 0.5}}$ & 10$_{\textcolor{gray}{\pm 1}}$ & 5$_{\textcolor{gray}{\pm 0}}$\\
Dreamer V3 & 9.7$_{\textcolor{gray}{\pm 0.9}}$ & 1.9$_{\textcolor{gray}{\pm 0.5}}$ & 44$_{\textcolor{gray}{\pm 2}}$ & 40$_{\textcolor{gray}{\pm 1}}$ & 7.2$_{\textcolor{gray}{\pm 0.8}}$ & 0.0$_{\textcolor{gray}{\pm 0.0}}$ & 45$_{\textcolor{gray}{\pm 2}}$ & 39$_{\textcolor{gray}{\pm 2}}$ & 9.9$_{\textcolor{gray}{\pm 0.7}}$ & \textbf{3.7}$_{\textcolor{gray}{\pm 0.3}}$ & 39$_{\textcolor{gray}{\pm 3}}$ & \textbf{33}$_{\textcolor{gray}{\pm 2}}$\\
TD-MPC2 & 13.8$_{\textcolor{gray}{\pm 1.4}}$ & 0.7$_{\textcolor{gray}{\pm 0.2}}$ & 32$_{\textcolor{gray}{\pm 1}}$ & 16$_{\textcolor{gray}{\pm 1}}$ & 9.5$_{\textcolor{gray}{\pm 1.0}}$ & 2.2$_{\textcolor{gray}{\pm 0.5}}$ & 40$_{\textcolor{gray}{\pm 4}}$ & 32$_{\textcolor{gray}{\pm 3}}$ & 9.2$_{\textcolor{gray}{\pm 1.3}}$ & 5.5$_{\textcolor{gray}{\pm 0.8}}$ & 35$_{\textcolor{gray}{\pm 2}}$ & 25$_{\textcolor{gray}{\pm 1}}$\\
\hline
Ours w/o MPFP & 12.3$_{\textcolor{gray}{\pm 1.2}}$ & 0.6$_{\textcolor{gray}{\pm 0.2}}$ & 46$_{\textcolor{gray}{\pm 3}}$ & 22$_{\textcolor{gray}{\pm 3}}$ & 8.6$_{\textcolor{gray}{\pm 1.1}}$ & 0.0$_{\textcolor{gray}{\pm 0.0}}$ & 45$_{\textcolor{gray}{\pm 2}}$ & 41$_{\textcolor{gray}{\pm 1}}$ & 10.5$_{\textcolor{gray}{\pm 1.6}}$ & 4.4$_{\textcolor{gray}{\pm 0.7}}$ & 35$_{\textcolor{gray}{\pm 3}}$ & 29$_{\textcolor{gray}{\pm 2}}$ \\
Ours & \textbf{8.7}$_{\textcolor{gray}{\pm 1.8}}$ & \textbf{0.5}$_{\textcolor{gray}{\pm 0.1}}$ & \textbf{61}$_{\textcolor{gray}{\pm 2}}$ & \textbf{42}$_{\textcolor{gray}{\pm 1}}$ & \textbf{4.3}$_{\textcolor{gray}{\pm 1.2}}$ & \textbf{0.0}$_{\textcolor{gray}{\pm 0.0}}$ & \textbf{70}$_{\textcolor{gray}{\pm 1}}$ & \textbf{60}$_{\textcolor{gray}{\pm 1}}$
& \textbf{6.3}$_{\textcolor{gray}{\pm 1.2}}$ & 6.9$_{\textcolor{gray}{\pm 1.0}}$ & \textbf{45}$_{\textcolor{gray}{\pm 3}}$ & 32$_{\textcolor{gray}{\pm 3}}$ \\
\hline
\end{tabular}}
\end{center}
\label{table3}
\vspace{-0.5cm}
\end{table*}

\begin{table} \small
\caption{Ablation studies on components of AES and MPFP in the challenging cross-room experimental configuration.}
\centering
\setlength{\tabcolsep}{1.7mm}{
\begin{tabular}{l c c c c}
\hline
\textbf{Ablation} & \textbf{AIKF}$\downarrow$ & \textbf{ABC}$\downarrow$ & \textbf{TCR}$\uparrow$ & \textbf{PSR}$\uparrow$ \\
\hline
w/o All AES & 16.7$_{\textcolor{gray}{\pm 1.3}}$ & 1.3$_{\textcolor{gray}{\pm 0.5}}$ & 10$_{\textcolor{gray}{\pm 2}}$ & 5$_{\textcolor{gray}{\pm 1}}$ \\
w/o AES-$p_{tc}$ & 10.1$_{\textcolor{gray}{\pm 1.0}}$ & 0.9$_{\textcolor{gray}{\pm 0.4}}$ & 51$_{\textcolor{gray}{\pm 3}}$ & 39$_{\textcolor{gray}{\pm 1}}$ \\
w/o AES-$p_{ies}$ & 12.7$_{\textcolor{gray}{\pm 1.1}}$ & 1.5$_{\textcolor{gray}{\pm 0.5}}$ & 34$_{\textcolor{gray}{\pm 4}}$ & 21$_{\textcolor{gray}{\pm 2}}$ \\
\hline
w/o MPFP in train & 12.8$_{\textcolor{gray}{\pm 1.4}}$ & 2.6$_{\textcolor{gray}{\pm 0.5}}$ & 31$_{\textcolor{gray}{\pm 1}}$ & 20$_{\textcolor{gray}{\pm 1}}$ \\
w/o MPFP in eval & 7.13$_{\textcolor{gray}{\pm 1.4}}$ & \textbf{0.8}$_{\textcolor{gray}{\pm 0.7}}$ & 68$_{\textcolor{gray}{\pm 4}}$ & 46$_{\textcolor{gray}{\pm 3}}$ \\
\hline
Our Full Method & \textbf{4.19}$_{\textcolor{gray}{\pm 1.2}}$ & 2.0$_{\textcolor{gray}{\pm 0.2}}$ & \textbf{81}$_{\textcolor{gray}{\pm 3}}$ & \textbf{54}$_{\textcolor{gray}{\pm 2}}$ \\
\hline
\end{tabular}}
\label{table4}
\vspace{-0.5cm}
\end{table}

\subsection{Comparative Studies}

To answer question (1), we first conduct comparative studies of the proposed method with all baselines in the challenging cross-room experimental configuration, and the experimental results are shown in Tab. \ref{table1}. Our approach significantly outperforms the SoTA EE-centric MM policy (N$^2$M$^2$), the RL-based maximum entropy policy (BHyRL), the world model-based methods (Dreamer V3 and TD-MPC2), and the auxiliary task distillation-based method (AuxDistill) in terms of reducing IK solving failures and increasing task success. As shown in Fig. \ref{fig4}, our method converges faster and better compared with strong baselines during training, reflecting that our method is more sample-efficient. The results of End-to-End MM indicate that it is difficult to learn MM policies based on whole-body control in an end-to-end manner using model-free RL. The results in Tab. \ref{table1} reflect that our method sometimes achieved sub-optimal ABC metrics. This is mainly because our approach reduces early task termination due to IK solving failures, and thus the extended episodes increase the likelihood of collisions. Furthermore, we train and evaluate our method and strong baselines in three other experimental configurations, with results shown in Tab. \ref{table2}. Either in static or dynamic scenes, our approach significantly reduces the number of IK solving failures (AIKF). In addition, our approach achieves competitive ABC metrics while significantly improving MM success rates (i.e., TCR and PSR). These results reflect the significant superiority of our method over the strong baselines. Training curves reflecting the sample efficiency can be found in the supplementary material. These advances are due to the fact that (1) AES enables our method paying more attention to those key state transitions that determine the success of the MM task; (2) RSSM-based MPFP helps to generate foresighted and collision-free mobile base motions to cooperate with the IK solving of the manipulator.

To answer question (2), both our method and three strong baselines trained in the cross-room experimental configuration are directly evaluated in the other three configurations without additional training. The results in Tab. \ref{table3} reflect that our approach can better generalize the RL-based MM agent to new spatial layouts, while N$^2$M$^2$ generalizes poorly. Compared to world model-based baselines, our approach is effective in reducing IK solving failures and collisions while improving task success. In addition, we find that RSSM-based MPFP is necessary because the spatial generalization of our method is significantly weaker when it is absent. The results in Tab. \ref{table2} and Tab. \ref{table3} also reflect that our method sometimes achieved sub-optimal ABC metrics. Such a phenomenon is especially prominent in dynamic scenes. Nevertheless, significant gains in the TCR metrics reflect the robustness and spatial generalization of our approach.

\subsection{Ablation Studies}

We answer questions (3) and (4) by respectively ablating different signals in AES and MPFP in the challenging cross-room experimental configuration. Based on the experimental results in Tab. \ref{table4}, we find that the absence of all AES signals significantly degrades the MM performance, especially by increasing IK solving failures and decreasing the TCR and PSR metrics. In addition, the absence of both informative experience selection (AES-$p_{ies}$) and task criticality (AES-$p_{tc}$) impairs MM performance, with AES-$p_{ies}$ having a greater impact. These results demonstrate the importance of AES for memorizing key experience fragments for robust MM. In addition, Fig. \ref{fig4} also reflects that our AES-based MM policy is sample-efficient. By ablating the RSSM-based MPFP in training and evaluation, respectively, we find that it always helps to facilitate the collaboration between the mobile base and the manipulator through farsighted dynamic imagination, which improves the MM performance. More studies on parameter $H$ and computational efficiency can be found in the supplementary material.

\subsection{Real-World Experiments}

We deploy our method to a real robotic platform consisting of a Husky A200 mobile base and a UR5 manipulator to verify its practicality. The robot's upper computer is a laptop with an Intel Core i9-13900HX CPU and GeForce RTX 4060 GPU. The robot uses a RealSense D435 RGB-D camera and YOLO v7 \cite{wang2023yolov7} to identify target objects. We use ROS to organize the hardware and software resources of the robotic system. Considering the security, we set the maximum speed and decision frequency of the mobile base to 0.3 $m/s$ and 1 Hz, respectively. Please see the supplementary material and videos for sim-to-real experiments.

\section{Conclusion and Limitations}

This paper investigates the problems of low sample efficiency and poor spatial generalization faced by MM in the context of embodied AI. Correspondingly, we propose the AES mechanism and RSSM-based MPFP to address these challenges. We find that AES, especially informative experience selection, can help RSSM memorize key experience fragments to significantly reinforce MM agents. RSSM-based prospective dynamic imagination can better generalize MM policies to new spatial layouts without additional training. Our problem modeling and experiments are carried out in an EE-centric MM framework, which fully combines the advantages of traditional robotic motion planning and robot learning. The practicality of our approach is verified by migrating it to a real-world robotic system without additional training.

\textbf{Limitations.} To focus on the contribution of the AES-enhanced RSSM to MM, we assumed an ideal observation space, especially using ground truth local occupancy maps directly. In practice, the construction of low-noise local occupancy maps from sensors (e.g., depth cameras and LiDAR) is essential.

\section*{Acknowledgments}
This work was supported in part by the Key Project of Xiangjiang Laboratory under 25XJ02003 and in part by the National Natural Science Foundation of China under 62272489, 62332020, and 62350004. This work was carried out in part using computing resources at the High-Performance Computing Center of Central South University.

\bibliographystyle{named}
\bibliography{ijcai26}

\end{document}